%% file: summary.tex
\documentclass[11pt,a4paper]{article}
\usepackage{bbm}
\usepackage[space]{grffile}
\usepackage{acl2015}
\usepackage{times}
\usepackage{url}
\usepackage{latexsym}
\usepackage{algpseudocode}
\usepackage{amsfonts}
\usepackage{amssymb}
\usepackage{amsthm}
\usepackage{algorithm}
\usepackage{amsmath}
\usepackage{booktabs}
\usepackage{multirow}
\usepackage{todonotes}
\usepackage{framed}
\usepackage{tikz}
\usepackage{subcaption}

\usetikzlibrary{positioning}

\newcommand{\enc}{\mathrm{enc}}

\newcommand{\xvec}{\mathbf{x}}
\newcommand{\yvec}{\mathbf{y}}

\newcommand{\mcY}{\mathcal{Y}}
\newcommand{\mcV}{\mathcal{V}}
\newcommand{\context}{\mathbf{y}_{\mathrm{c}}}
\newcommand{\embcontext}{\mathbf{\tilde{y}}_{\mathrm{c}}}
\newcommand{\inpcontext}{\mathbf{\tilde{x}}}

\newcommand{\Uvec}{\mathbf{U}}
\newcommand{\Evec}{\mathbf{E}}
\newcommand{\Gvec}{\mathbf{G}}
\newcommand{\Fvec}{\mathbf{F}}
\newcommand{\Pvec}{\mathbf{P}}
\newcommand{\pvec}{\mathbf{p}}
\newcommand{\Vvec}{\mathbf{V}}
\newcommand{\Wvec}{\mathbf{W}}
\newcommand{\hvec}{\mathbf{h}}
\newcommand{\reals}{\mathbb{R}}
\renewcommand{\algorithmicrequire}{\textbf{Input:}}
\renewcommand{\algorithmicensure}{\textbf{Output:}}
\def\argmax{\operatornamewithlimits{arg\,max}}
\def\kargmax{\operatornamewithlimits{K-arg\,max}}

\algnewcommand{\LineComment}[1]{\State \(\triangleright\) #1}
\title{A Neural Attention Model for
Abstractive Sentence Summarization}

\author{Alexander M. Rush \\
  Facebook AI Research / \\
  Harvard SEAS \\
  srush@seas.harvard.edu \\
  \And Sumit Chopra \\
  Facebook AI Research\\
  spchopra@fb.com
\And Jason Weston \\
Facebook AI Research\\
jase@fb.com}

\date{}

\begin{document}
\maketitle

\begin{abstract}
  Summarization based on text extraction is inherently limited, but
  generation-style abstractive methods have proven challenging to
  build. In this work, we propose a fully data-driven approach to
  abstractive sentence summarization. Our method utilizes a local
  attention-based model that generates each word of the summary
  conditioned on the input sentence. While
  the model is structurally simple, it can easily be trained
  end-to-end and scales to a large amount of training data. The model
  shows significant performance gains on the DUC-2004 shared task
  compared with several strong baselines.
\end{abstract}

\input{introduction}

\input{background}

\input{model}

\input{generation}

\input{related}

\input{methods}

\input{experiments}

\bibliographystyle{acl}
\bibliography{full}

\end{document}

%% file: introduction.tex
\section{Introduction}

Summarization is an important challenge of natural language
understanding.  The aim is to produce a condensed representation of an
input text that captures the core meaning of the original. Most
successful summarization systems utilize \textit{extractive}
approaches that crop out and stitch together portions of
the text to produce a condensed version. In contrast,
\textit{abstractive} summarization attempts to produce a bottom-up summary, aspects of which may not appear as part of the
original.

\begin{figure}
  \centering
  \vspace*{-0.5cm}
  \includegraphics[width=1.0\columnwidth]{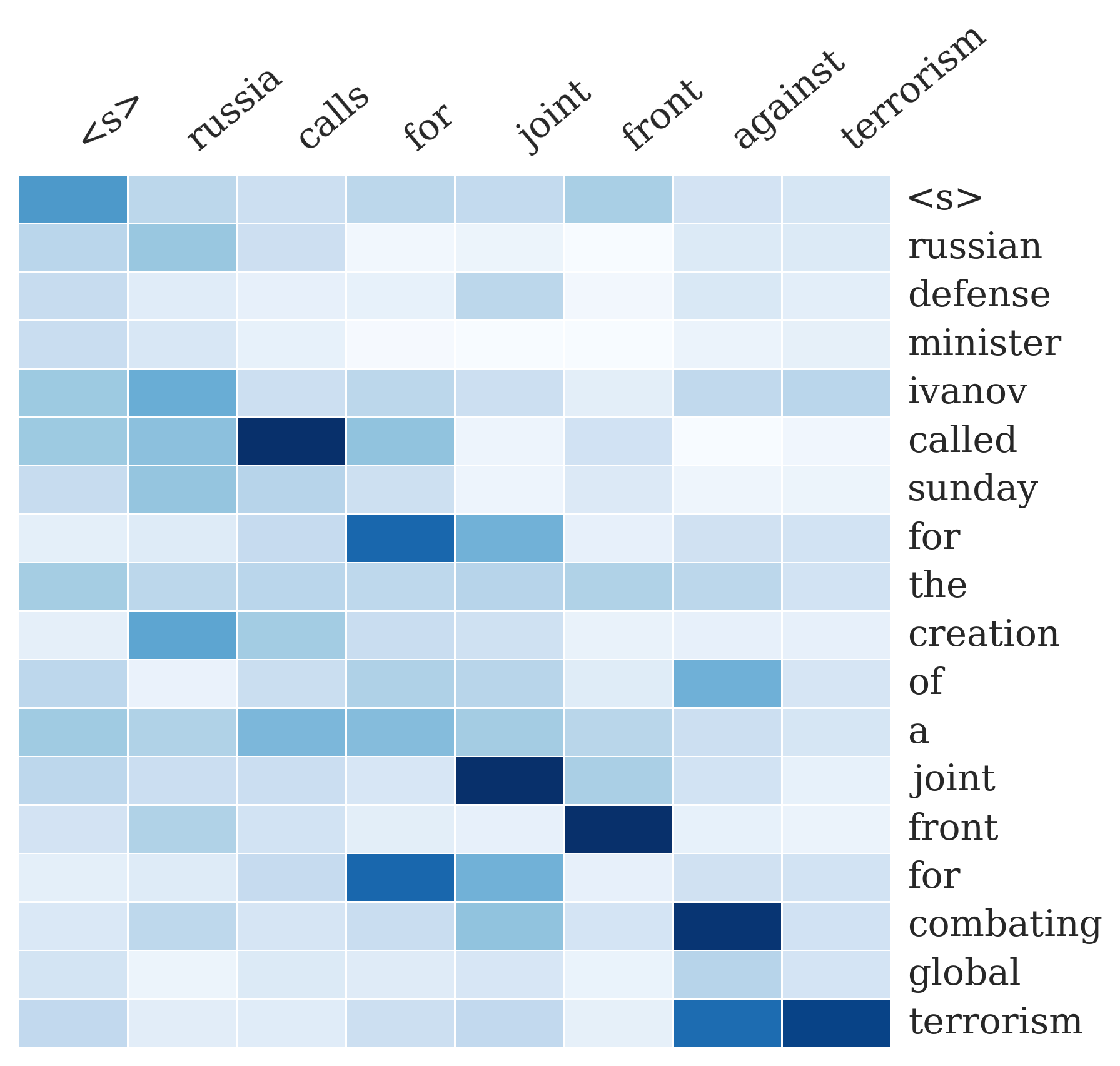}
  \caption{ \small \label{fig:aligna} Example output of the attention-based summarization (ABS) system. The heatmap represents a soft alignment between the input (right) and the generated summary (top). The columns represent the distribution over the input after generating each word.}
\end{figure}

\begin{figure*}[t!]
  \small 
  \begin{framed}
    
    Input $(\xvec_1, \ldots, \xvec_{18})$. First sentence of article:
    
    russian defense minister ivanov called sunday for the creation of a joint front for combating global terrorism
    \vspace{0.1cm}
    
    Output $(\yvec_1, \ldots, \yvec_8)$. Generated headline:

    \textit{russia calls for joint front against \textbf{terrorism}}  \hspace*{0.5cm} $ \Leftarrow$ \hspace*{0.5cm} $g(\mathrm{\textit{terrorism}}, \xvec, \mathrm{\textit{for, joint, front, against}})$  
  \end{framed}
  \caption{\small \label{fig:sumex} Example input sentence and the generated summary. The score of generating $\yvec_{i+1}$
    (\texttt{terrorism}) is based on the context $\context$
    (\texttt{for} $\ldots$ \texttt{against}) as well as the input $\xvec_1 \ldots \xvec_{18}$. 
    Note that the summary generated is abstractive which makes it possible to \textit{generalize} (\texttt{russian defense minister} to \texttt{russia}) and \textit{paraphrase} (\texttt{for combating} to \texttt{against}), in addition to \textit{compressing}  (dropping \texttt{the creation of}), see \newcite{jing2002using} for a survey of these editing operations.}
\end{figure*}

We focus on the task of sentence-level summarization.
While much work on this task has looked at deletion-based sentence
compression techniques (\newcite{knight2002summarization}, among many others), studies of
human summarizers show that it is common to apply various other
operations while condensing, such as paraphrasing, generalization, and reordering  \cite{jing2002using}. Past work has
modeled this abstractive summarization problem either using
linguistically-inspired constraints \cite{dorr2003hedge,zajic2004bbn}
or with syntactic transformations of the
input text \cite{cohn2008sentence,woodsend2010generation}.
These approaches are described in more detail in Section~\ref{sec:related}.

We instead explore a fully data-driven approach for generating
abstractive summaries. Inspired by the recent success
 of neural machine translation, we combine a neural
language model with a contextual input encoder. Our encoder
is modeled off of the attention-based encoder of
\newcite{bahdanau2014neural} in that it learns a latent soft alignment
over the input text to help inform the summary (as shown in Figure~\ref{fig:aligna}). Crucially both the
encoder and the generation model are trained jointly on the sentence
summarization task.  The model is described in detail in
Section~\ref{sec:model}. Our model also incorporates
a beam-search decoder as well as additional features to model extractive elements; these aspects are discussed in
Sections~\ref{sec:additional} and~\ref{sec:tuning}.

This approach to summarization, which we call
Attention-Based Summarization (\textsc{Abs}), incorporates less linguistic structure than
comparable abstractive summarization approaches, but can easily scale to train on a
large amount of data. Since our system makes no assumptions about the
vocabulary of the generated summary it can be trained directly on any
document-summary pair.\footnote{In contrast to a large-scale sentence
compression systems like \newcite{filippova2013overcoming} which
require monotonic aligned compressions.} This allows us to train a
summarization model for headline-generation on a corpus of article
pairs from Gigaword \cite{graff2003english} consisting of around 4
million articles. An example of generation is given in Figure~\ref{fig:sumex}, and we discuss the details of this task in Section~\ref{sec:methods}.

To test the effectiveness of this approach we run extensive
comparisons with multiple abstractive and extractive baselines,
including traditional syntax-based systems, integer linear
program-constrained systems, information-retrieval style approaches,
as well as statistical phrase-based machine
translation. Section~\ref{sec:results} describes the results of these
experiments. Our approach outperforms a machine translation system
trained on the same large-scale dataset and yields a large
improvement over the highest scoring system in the DUC-2004
competition.

%% file: background.tex
\section{Background}
\label{sec:background}

We begin by defining the sentence summarization
task. Given
an input sentence, the goal is to produce a condensed summary. Let the
input consist of a sequence of $M$ words $\xvec_1, \ldots, \xvec_M$
coming from a fixed vocabulary $\mcV$ of size $|\mcV| = V$ . We
will represent each word as an indicator vector $\xvec_i \in \{0,
1\}^{V}$ for $i \in \{1,\ldots, M\}$, sentences as a sequence of
indicators, and $\mathcal{X}$ as the set of possible inputs.  Furthermore define the notation $\xvec_{[i,j,k]}$ to
indicate the sub-sequence of elements $i,j,k$.

A summarizer takes $\xvec$ as input and outputs a shortened sentence $\yvec$ of length $N
< M$. We will assume that the words in
the summary also come from the same vocabulary $\mcV$ and that the output is a sequence
$\yvec_1, \ldots, \yvec_N$. Note that in contrast to related tasks, like machine translation, we will  assume
that the output length $N$ is fixed, and that the system knows the length of the
summary before generation.\footnote{For the DUC-2004 evaluation, it is actually the number of \textit{bytes} of the output that is capped. More detail is given in Section~\ref{sec:methods}.}

Next consider the problem of generating summaries. Define the set $\mcY
\subset (\{0,1\}^{V}, \ldots, \{0,1\}^{V})$ as all possible sentences of
length $N$, i.e. for all $i$ and $\yvec \in \mcY$, $\yvec_i$ is an
indicator.  We say a system is \textit{abstractive} if it tries to
find the optimal sequence from this set $\mcY$, 
\begin{eqnarray} \argmax_{\yvec \in \mcY} s(\xvec, \yvec), \end{eqnarray}
\noindent under a scoring function $s: \mathcal{X} \times
\mathcal{Y} \mapsto \reals$. Contrast this to a fully
\textit{extractive} sentence summary\footnote{Unfortunately the literature is inconsistent on the formal definition of this distinction. Some systems self-described as abstractive would be extractive under our definition. } which transfers words from the input:
\begin{eqnarray}\argmax_{m \in \{1,\ldots M\}^N } s(\xvec, \xvec_{[m_1, \ldots,m_N]}), \label{eq:ext}\end{eqnarray}  
\noindent or to the related problem of sentence \textit{compression} that concentrates on deleting words from the input: 
\begin{eqnarray}
\argmax_{m \in \{1,\ldots M\}^N, m_{i-1} < m_{i} } s(\xvec, \xvec_{[m_1, \ldots,m_N]}). \end{eqnarray}
While abstractive summarization poses a more difficult generation challenge, the lack of hard constraints gives the system more freedom in generation and allows it to fit with a wider range of training data.

In this work we focus on factored scoring functions, $s$, that take into account a fixed window of previous words:
\begin{eqnarray} s(\xvec, \yvec) \approx \sum_{i=0}^{N-1} g(\yvec_{i+1}, \xvec, \context), \label{eq:factor}
\end{eqnarray}
\noindent where we define $\context \triangleq \yvec_{[i-C+1, \ldots, i]}$ for a window of size $C$. 

In particular consider the conditional log-probability of a summary given the input, $s(\xvec, \yvec) = \log p(\yvec | \xvec; \theta)$. We can write this as:
\begin{eqnarray*} 
  \log p(\yvec | \xvec; \theta) &\approx&\sum_{i=0}^{N-1} \log p(\yvec_{i+1} | \xvec, \context; \theta), 
\end{eqnarray*}

\noindent where we make a Markov assumption on the length of the
context as size $C$ and assume for $i < 1$, $\yvec_i$  is a special start symbol $\langle S \rangle$. 

With this scoring function in mind, our main focus will
be on modelling the local conditional distribution: $p(\yvec_{i+1} |
\xvec, \context; \theta)$. The next section defines a parameterization for
this distribution,
in Section~\ref{sec:additional}, we return to
the question of generation for factored models, and 
in Section~\ref{sec:tuning} we introduce
a modified factored scoring function.

%% file: model.tex
\section{Model}
\label{sec:model}

The distribution of interest, $p(\yvec_{i+1} | \xvec, \context ; \theta)$, is  a conditional language model based on the input
sentence $\xvec$. Past work on summarization and compression has used
a noisy-channel approach to split and independently estimate a 
language model and a conditional summarization model \cite{banko2000headline,knight2002summarization,daume2002noisy},
i.e.,
\[ \argmax_{\yvec} \log p(\yvec | \xvec) = \argmax_{\yvec} \log p(\yvec) p(\xvec | \yvec) \]

\noindent where $p(y)$ and $p(x|y)$ are estimated separately.  Here we
instead follow work in neural machine translation and directly
parameterize the original distribution as a neural network. The
network contains both a neural probabilistic language model and an
encoder which acts as a conditional summarization model.

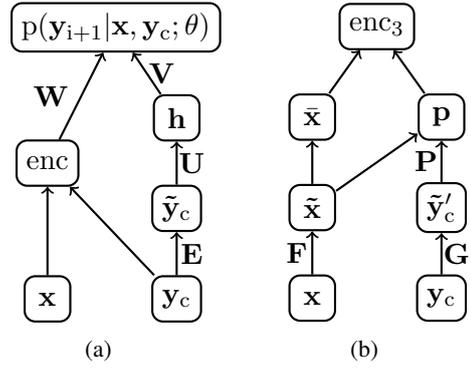
\begin{figure}
  \centering

  \begin{subfigure}{0.3\columnwidth}

  \begin{tikzpicture}[node distance=1.2cm]
    \tikzstyle{box} = [draw, thick, rounded corners, minimum size=0.6cm];
    \tikzstyle{line} = [draw, thick];
    \node(article)[box]{$\xvec$};
    \node(context)[box, right of=article, xshift =0.5cm]{$\context$ };
    \node(enc)[box, yshift=0.6cm, above of=article]{$\mathrm{enc}$};
    \path[line, ->] (article) -- (enc);
    \path[line, ->] (context) -- (enc);

    \node(embcontext)[box, above of = context] {$\embcontext$};

    \node (P) [box, above of = embcontext] {$\hvec$}; 

    \node (out) [box, above of=P, xshift=-0.8cm] {$\mathrm{p(\yvec_{i+1} | \xvec, \context;\theta)}$} ;

    \path[line, ->] (enc) --node[xshift=-0.4cm]{$\Wvec$} (out);

    \path[line, draw, ->](context) -- node[xshift=0.2cm]{$\Evec$}  (embcontext) ;
    \path[line, draw, ->](embcontext) -- node[xshift=0.2cm]{$\Uvec$}  (P) ;

    \path[line, draw, ->](P) --node[xshift=0.2cm] {$\Vvec$}   (out) ;

  \end{tikzpicture}
  \caption{\label{fig:schdecoder}}
  \end{subfigure}
  \hspace{1cm}
  \begin{subfigure}{0.3\columnwidth}
  \begin{tikzpicture}[node distance=1.2cm]
    \tikzstyle{box} = [draw, thick, rounded corners, minimum size=0.6cm];
    \tikzstyle{line} = [draw, thick];
    \node(article)[box]{$\xvec$};
    \node(context)[box, right of=article, xshift =0.5cm]{$\context$ };
    \node(inpemb)[box, above of=article]{$\inpcontext$};
    \node(embcontext)[box, above of = context] {$\embcontext'$};
    \node (Q) [box, above of = inpemb] {$\bar{\mathbf{x}}$}; 
    \node (P) [box, above of = embcontext] {$\pvec$}; 

    \path (Q) -- node (out) [box, yshift=1.2cm] {$\mathrm{enc}_3$} (P);

    \path[line, draw, ->](article) -> node[xshift=-0.2cm]{$\Fvec$} (inpemb);
    \path[line, draw, ->] (inpemb) ->node[xshift=-0.2cm]{} (Q);
    \path[line, draw, ->](context) -- node[xshift=0.2cm]{$\Gvec$}  (embcontext) ;
    \path[line, draw, ->](embcontext) -- node[xshift=-0.2cm]{$\Pvec$}  (P) ;
    \path[line, draw, ->](inpemb) --   (P) ;

    \path[line, draw, ->](P) --   (out) ;
    \path[line, draw, ->](Q) --   (out) ;

  \end{tikzpicture}
  \caption{\label{fig:schencoder}}
  \end{subfigure}

  \caption{\small \label{fig:decoder} (a) A network diagram for the NNLM decoder with additional encoder element. (b)
    A network diagram for the attention-based encoder $\mathrm{enc}_3$.  
  }
\end{figure}

\subsection{Neural Language Model}

The core of our parameterization is a language model for estimating
the contextual probability of the next word. The language model is
adapted from a standard feed-forward neural network language model
(NNLM), particularly the class of NNLMs described by
\newcite{bengio2003neural}. The full model is:
\begin{eqnarray*}
 p(\yvec_{i+1} | \context, \xvec; \theta) &\propto& \exp(\Vvec \hvec + \Wvec \enc(\xvec, \context)), \\  
  \embcontext &=& [\Evec \yvec_{i-C + 1}, \dots,\Evec \yvec_{i}],  \\ 
 \hvec &=& \tanh(\Uvec  \embcontext ).  
\end{eqnarray*}
\noindent The parameters are $\theta = (\Evec, \Uvec , \Vvec,
\Wvec)$ where $\Evec \in \reals^{D \times V}$ is a word embedding
matrix, $\Uvec \in \reals^{(C D) \times H} $, $\Vvec \in \reals^{V
  \times H}$, $\Wvec \in \reals^{V \times H}$ are weight
matrices,\footnote{Each of the weight matrices $\Uvec$, $\Vvec$,
  $\Wvec$ also has a corresponding bias term. For readability, we
  omit these terms throughout the paper.} $D$ is the size of the word
embeddings, and $\hvec$ is a hidden layer of size $H$. The black-box function $\enc$
is a contextual encoder term that returns a vector of size $H$
representing the input and current context; we consider several
possible variants, described subsequently. Figure~\ref{fig:schdecoder}
gives a schematic representation of the decoder
architecture.

\subsection{Encoders}

Note that without the encoder term this represents a standard language
model. By incorporating in $\enc$ and training the two elements
jointly we crucially can incorporate the input text into
generation. We discuss next several possible instantiations of the
encoder.

\paragraph{Bag-of-Words Encoder}

Our most basic model simply uses the bag-of-words of the input
sentence embedded down to size $H$, while ignoring properties of the
original order or relationships between neighboring words. We write
this model as:
\begin{eqnarray*}
  \enc_1(\xvec, \context) &=&  \mathbf{p}^\top \inpcontext,  \\ 
  \mathbf{p} & =  & [1/M, \ldots, 1/M],   \\ 
  \inpcontext &=& [\Fvec \xvec_1, \dots, \Fvec \xvec_{M}].
\end{eqnarray*}
\noindent Where the input-side embedding matrix $\Fvec \in \reals^{H
  \times V}$ is the only new parameter of the encoder 
 and $\mathbf{p} \in [0,1]^{M}$ is a
uniform distribution over the input words.

For summarization this model can
capture the relative importance of words to distinguish content words
from stop words or embellishments. Potentially the model can also
learn to combine words; although it is inherently limited in
representing contiguous phrases.

\paragraph{Convolutional Encoder}

To address some of the modelling issues with bag-of-words
we also consider using a deep convolutional encoder for the input
sentence. This architecture improves on the bag-of-words model by
allowing local interactions between words while also not requiring the
context $\context$ while encoding the input.

We utilize a standard time-delay neural network (TDNN) architecture,
alternating between temporal convolution layers and max pooling
layers.

\begin{small}
\begin{eqnarray}
\forall j, \;\; \enc_2(\xvec, \context)_j &=& \max_{i} \tilde{\xvec}^{L}_{i,j} , \label{eq:max}  \\
 \forall i, l\in \{1,\ldots L\}, \;\; \tilde{\xvec}_{j}^{l} &=& \tanh ( \max \{ \bar{\xvec}^{l}_{2i-1}, \bar{\xvec}^{l}_{2i} \} ), \nonumber \\ \label{eq:pool} \\ 
 \forall i, l \in \{1,\ldots L\}, \;\; \bar{\xvec}_i^{l} &=&  \mathbf{Q}^{l} \inpcontext^{l-1}_{[i-Q, \ldots, i +Q]} , \label{eq:conv} \\ 
 \inpcontext^{0} &=& [\Fvec \xvec_1, \dots, \Fvec \xvec_{M}].  
\end{eqnarray}
\end{small}
Where $\mathbf{F}$ is a word embedding matrix and $\mathbf{Q}^{L\times
  H \times 2Q +1 }$ consists of a set of filters for each layer $\{1,
\ldots L\}$.  Eq.~\ref{eq:conv} is a temporal (1D) convolution layer,
Eq.~\ref{eq:pool} consists of a 2-element temporal max pooling layer
and a pointwise non-linearity, and final output Eq.~\ref{eq:max} is a
max over time. At each layer $\tilde{x}$ is one half the size of
$\bar{x}$. For simplicity we assume that the convolution is padded at
the boundaries, and that $M$ is greater than $2^L$ so that the
dimensions are well-defined. 

\paragraph{Attention-Based Encoder}

While the convolutional encoder has richer capacity than bag-of-words, it
still is required to produce a single representation for the entire
input sentence. A similar issue in machine translation inspired 
\newcite{bahdanau2014neural} to instead utilize an attention-based 
contextual encoder that constructs a representation based on the 
generation context. Here we note that if we exploit this context, we 
can actually use a rather simple model similar to bag-of-words:
\begin{eqnarray*}
 \enc_3(\xvec, \context) &=& \mathbf{p}^\top \bar{\xvec},    \\
 \mathbf{p} & \propto  & \exp(\inpcontext  \Pvec \embcontext' ),   \\ 
 \inpcontext &=& [\Fvec \xvec_1, \dots, \Fvec \xvec_{M}],  \\ 
  \mathbf{\embcontext'} &=& [\Gvec \yvec_{i-C + 1}, \dots,\Gvec \yvec_{i}], \\ 
 \forall i \;\;\; \bar{\xvec}_i &=& \sum_{q=i-Q}^{i+Q}  \inpcontext_{i} / Q.
\end{eqnarray*}
\noindent Where $\Gvec \in \reals^{D \times V}$ is an embedding of the
context, $\Pvec \in \reals^{H \times (C D)}$ is a new weight matrix parameter mapping
between the context embedding and input embedding, and $Q$ is a
smoothing window. The full model is shown in Figure~\ref{fig:schencoder}.

Informally we can think of this model as simply replacing the uniform
distribution in bag-of-words with a learned soft alignment,
$\mathbf{P}$, between the input and the summary.
Figure~\ref{fig:aligna} shows an example of this distribution
$\mathbf{p}$ as a summary is generated. The soft alignment is then used to
weight the smoothed version of the input $\bar{\xvec}$ when
constructing the representation. For instance if the current context
aligns well with position $i$ then the words $\xvec_{i-Q}, \ldots,  \xvec_{i+Q}$
are highly weighted by the encoder.
Together with the NNLM, this model can be seen as a stripped-down version of the
attention-based neural machine translation model.\footnote{To be explicit, compared to \newcite{bahdanau2014neural} our model uses
  an NNLM instead of a target-side LSTM, source-side windowed
  averaging instead of a source-side bi-directional RNN, and a
  weighted dot-product for alignment instead of an alignment MLP.
}

\subsection{Training}
\label{sec:train}
The lack of generation constraints makes it possible to train the
model on arbitrary input-output pairs.  Once we have defined
the local conditional model, $p(\yvec_{i+1}| \xvec, \context ; \theta)$,  we can estimate the parameters to minimize the
negative log-likelihood of a set of summaries. Define this training
set as consisting of $J$ input-summary pairs $(\xvec^{(1)},
\yvec^{(1)}), \ldots, (\xvec^{(J)}, \yvec^{(J)})$.  The negative
log-likelihood conveniently factors\footnote{This is dependent on using the gold standard contexts $\context$. An alternative is to use the predicted context within a structured or reenforcement-learning style objective.} into a term for each token
in the summary:

\begin{small}
\begin{eqnarray*} 
\mathrm{NLL}(\theta) &=& -\sum_{j=1}^J \log p(\yvec^{(j)}| \xvec^{(j)};\theta), \\
&=& -\sum_{j=1}^J \sum_{i=1}^{N-1} \log p(\yvec_{i+1}^{(j)}| \xvec^{(j)}, \context ; \theta). 
\end{eqnarray*}
\end{small}

\noindent We minimize NLL by using mini-batch stochastic gradient
descent. The details are described further in
Section~\ref{sec:methods}.

%% file: generation.tex
\section{Generating Summaries}
\label{sec:additional}

We now return to the problem of generating summaries. Recall from Eq.~\ref{eq:factor} that our goal is to find,  
\begin{eqnarray*}
\yvec^* &=& \argmax_{\yvec \in \mcY} \sum_{i=0}^{N-1} g(\yvec_{i+1}, \xvec, \context).
\end{eqnarray*}

\noindent Unlike phrase-based machine translation where inference is
NP-hard, it actually is tractable in theory to compute $\yvec^*$.
Since there is no explicit hard alignment constraint, Viterbi decoding
can be applied and requires $O(N V^C)$ time to find an exact
solution. In practice though $V$ is large enough to make this
difficult. An alternative approach is to approximate the $\argmax$ with a strictly
greedy or deterministic decoder. 

A compromise between exact and greedy decoding is to use a beam-search
decoder (Algorithm~\ref{fig:alg}) which maintains the full vocabulary $\mcV$ while limiting itself
to $K$ potential hypotheses at each position of the summary. This has been the standard
approach for neural MT models \cite{bahdanau2014neural,sutskever2014sequence,luong2014addressing}. The beam-search algorithm is shown here, modified for the feed-forward model:

\begin{algorithm}[ht]
  \small
  \begin{algorithmic}
    \Require{Parameters $\theta$, beam size $K$, input $\xvec$}
    \Ensure{Approx. $K$-best summaries }
    \State{$\pi[0] \gets \{ \epsilon \}$}
    \State{$\mathcal{S} =   \mcV \mathrm{\ if \ abstractive\ else \ } \{ \xvec_i\  | \ \forall i \}$}
    \For{$i = 0$ to $ N-1$}
    \LineComment{Generate Hypotheses}

    \State{
      $\mathcal{N} \gets \left\{ \begin{matrix} [\yvec, \yvec_{i+1}]  &|&  \yvec  \in \pi[i],  \yvec_{i+1} \in \mathcal{S} 
        \end{matrix} \right\} $}    
    \State{\LineComment{Hypothesis Recombination}}
    \State{$\mathcal{H} \gets \left\{ \begin{matrix} \yvec \in \mathcal{N} & | &  s(\yvec , \xvec) > s(\yvec', \xvec) \\
          && \forall \yvec' \in \mathcal{N} \mathrm{\ s.t.\ } \context = \context' \end{matrix} \right\} $ }
    \State{\LineComment{Filter K-Max}}
    \State{$\displaystyle \pi[i+1] \gets \kargmax_{\yvec \in \mathcal{H}}  g(\yvec_{i+1}, \context, \xvec) + s(\yvec, \xvec) $}
    \EndFor
    \State{\Return{ $\pi[N]$}}
  \end{algorithmic}
  \caption{\label{fig:alg} Beam Search }
\end{algorithm}

As with Viterbi this beam search algorithm is much simpler than beam
search for phrase-based MT. Because there is no explicit constraint
that each source word be used exactly once there is no need to
maintain a bit set and we can simply move from left-to-right generating
words. The beam search algorithm requires $O(KNV)$ time. From a
computational perspective though, each round of beam search is
dominated by computing $p(\yvec_i | \xvec, \context)$ for each of the
$K$ hypotheses. These can be computed as a mini-batch, which 
in practice greatly reduces the factor of $K$.

\section{Extension: Extractive Tuning}
\label{sec:tuning}

While we will see that the attention-based model is effective at
generating summaries, it does miss an important aspect seen in
the human-generated references. In particular the abstractive model
does not have the capacity to find extractive word matches when
necessary, for example transferring unseen proper noun phrases from the
input. Similar issues have also been observed in neural
translation models particularly in terms of translating rare words
\cite{luong2014addressing}.

To address this issue we experiment with tuning a very small set of
additional features that trade-off the abstractive/extractive 
tendency of the system. We do this by modifying our scoring function to 
directly estimate the probability of a summary using a log-linear model,
as is standard in machine translation:
\begin{eqnarray*} 
p(\yvec | \xvec;\theta, \alpha ) &\propto& \exp( \mathbf{\alpha}^\top \sum_{i = 0}^{N-1} f(\yvec_{i+1}, \xvec, \context) ).
\end{eqnarray*}

\noindent Where $\alpha \in \reals^5$ is a weight vector and $f$ is a feature function. Finding the best summary under this distribution corresponds to maximizing a factored scoring function $s$,
\begin{eqnarray*} 
 s(\yvec, \xvec) &=&   \sum_{i = 0}^{N-1} \mathbf{\alpha}^\top  f(\yvec_{i+1}, \xvec, \context) . 
\end{eqnarray*}
\noindent where $g(\yvec_{i+1}, \xvec, \context) \triangleq \mathbf{\alpha}^\top  f(\yvec_{i+1}, \xvec, \context)$ to satisfy Eq.~\ref{eq:factor}.
The function $f$ is defined to combine the local conditional probability with some additional indicator featrues:  
\begin{align*}
f(&\yvec_{i+1}, \xvec, \context)  = [\ \log p(\yvec_{i+1} | \xvec, \context; \theta),  \\
 &\mathbbm{1}\{\exists j.\ \yvec_{i+1} = \xvec_{j}\ \}, \\
 &\mathbbm{1}\{\exists j.\ \yvec_{i+1-k} = \xvec_{j-k}\ \forall k\in\{0,1\}\},  \\
 &\mathbbm{1}\{\exists j.\ \yvec_{i+1-k} = \xvec_{j-k}\ \forall k\in\{0,1,2\} \}, \\
 &\mathbbm{1}\{\exists k>j.\ \yvec_{i} = \xvec_{k}, \yvec_{i+1} = \xvec_j\}\ ].
\end{align*}

\noindent These features correspond to indicators of unigram, bigram, and trigram match with the input as well as reordering of input words. Note that setting $\alpha = \langle 1, 0, \ldots, 0\rangle$ gives a model identical to standard \textsc{Abs}.

After training the main neural model, we fix $\theta$ and tune the $\alpha$ parameters. We
follow the statistical machine translation setup and use minimum-error
rate training (MERT) to tune for the summarization metric on tuning
data \cite{och2003minimum}. This tuning step is also identical to
the one used for the phrase-based machine translation baseline.

%% file: related.tex
\section{Related Work}
\label{sec:related}

Abstractive sentence summarization has been traditionally connected to
the task of headline generation.  Our work is similar to early work of
\newcite{banko2000headline} who developed a statistical machine
translation-inspired approach for this task using a corpus of
headline-article pairs. We extend this approach by: (1) using a neural
summarization model as opposed to a count-based noisy-channel model, (2)
training the model on much larger scale (25K compared to 4 million
articles), (3) and allowing fully abstractive decoding. 

This task was standardized around the DUC-2003 and DUC-2004
competitions \cite{over2007duc}. The \textsc{Topiary} system
\cite{zajic2004bbn} performed the best in this task, and is described in
detail in the next section.  We point interested readers to the
DUC web page (\url{http://duc.nist.gov/}) for the full list of systems
entered in this shared task.

More recently, \newcite{cohn2008sentence} give a compression method
which allows for more arbitrary transformations. They extract tree
transduction rules from aligned, parsed texts and learn weights on
transfomations using a max-margin learning
algorithm. \newcite{woodsend2010generation} propose a
quasi-synchronous grammar approach utilizing both
context-free parses and dependency parses to produce legible
summaries. Both of these approaches differ from ours in that they
directly use the syntax of the input/output sentences.  The latter
system is \textsc{W\&L} in our results; we attempted to train the
former system \textsc{T3} on this dataset but could not train it at
scale.

In addition to \newcite{banko2000headline} there has been some work
using  statistical machine translation directly for abstractive
summary. \newcite{wubben2012sentence} utilize \textsc{Moses} directly
as a method for text simplification.

Recently \newcite{filippova2013overcoming} developed a strictly
extractive system that is trained on a relatively large corpora (250K
sentences) of article-title pairs. Because their
focus is extractive compression, the sentences are
transformed by a series of heuristics such that the words are in
monotonic alignment. Our system does not require this alignment step
but instead uses the text directly. 

\paragraph{Neural MT}

This work is closely related to recent work on neural network language
models (NNLM) and to work on neural machine translation. The core of
our model is a NNLM based on that of \newcite{bengio2003neural}.

Recently, there have been several papers about models for machine
translation
\cite{kalchbrenner2013recurrent,cho2014learning,sutskever2014sequence}. Of
these our model is most closely related to the attention-based model
of \newcite{bahdanau2014neural}, which explicitly finds a soft
alignment between the current position and the input source. Most of
these models utilize recurrent neural networks (RNNs) for generation
as opposed to feed-forward models. We hope to incorporate an RNN-LM 
in future work.

%% file: methods.tex
\section{Experimental Setup}
\label{sec:methods}

We experiment with our attention-based sentence summarization model on the task
of headline generation. In this section we describe the
corpora used for this task, the baseline methods we compare with, and
implementation details of our approach.

\subsection{Data Set}

The standard sentence summarization evaluation set is associated with
the DUC-2003 and DUC-2004 shared tasks \cite{over2007duc}. The data
for this task consists of 500 news articles from the New York Times
and Associated Press Wire services each paired with 4 different
human-generated reference summaries (not actually headlines), capped at 75 bytes. This data set
is evaluation-only, although the similarly sized DUC-2003 data
set was made available for the task. The expectation is for a
summary of roughly 14 words, based on the text of a complete article (although we
only make use of the first sentence).  The full data set is available
by request at \url{http://duc.nist.gov/data.html}.

For this shared task, systems were entered and evaluated using
several variants of the recall-oriented ROUGE metric \cite{lin2004rouge}. To make
recall-only evaluation unbiased to length, output of all systems is
cut-off after 75-characters and no bonus is given for shorter
summaries. Unlike BLEU which interpolates various n-gram matches,
there are several versions of ROUGE for different match lengths. The
DUC evaluation uses ROUGE-1 (unigrams), ROUGE-2 (bigrams), and ROUGE-L
(longest-common substring), all of which we report.

In addition to the standard DUC-2014 evaluation, we also report
evaluation on single reference headline-generation using a randomly
held-out subset of Gigaword.  This evaluation is closer to the task
the model is trained for, and it allows us to use a bigger evaluation
set, which we will include in our code release. For this evaluation,
we tune systems to generate output of the average title length.

For training data for both tasks, we utilize the annotated Gigaword data set
\cite{graff2003english,napoles2012annotated}, which consists of
standard Gigaword, preprocessed with Stanford CoreNLP tools
\cite{manning2014stanford}. Our model only uses annotations for
tokenization and sentence separation, although several of the
baselines use parsing and tagging as well. Gigaword contains around
9.5 million news articles sourced from various domestic and
international news services over the last two decades.

For our training set, we pair the headline of each article with its
first sentence to create an input-summary pair. While the model could
in theory be trained on any pair, Gigaword contains many spurious
headline-article pairs.  We therefore prune training based on the following
heuristic filters: (1) Are there no non-stop-words in common? (2)
Does the title contain a byline or other extraneous editing marks? (3)
Does the title have a question mark or colon? After applying these
filters, the training set consists of roughly $J=4$ million
title-article pairs. We apply a minimal preprocessing step using 
PTB tokenization, lower-casing, replacing all digit characters with
\#, and replacing of word types seen less than 5 times with UNK.  We
also remove all articles from the time-period of the DUC evaluation. 
release.

The complete input training vocabulary consists of $119$ million
word tokens and 110K unique word types with an average sentence size of
$31.3$ words.  The headline vocabulary consists of $31$ million tokens
and 69K word types with the average title of length $8.3$ words (note that this is significantly shorter than the DUC summaries). On average
there are $4.6$ overlapping word types between the headline and the
input; although only $2.6$ in the first 75-characters of the input.

\begin{table*}[ht!]
  \centering
  \footnotesize
  \begin{tabular}{lccccccccc}
    \toprule
                      & \multicolumn{3}{c}{DUC-2004 } && \multicolumn{4}{c}{Gigaword} \\
    Model             & ROUGE-1 & ROUGE-2 & ROUGE-L && ROUGE-1 & ROUGE-2 & ROUGE-L & Ext. \%  \\ 
    \midrule
    \textsc{IR}           & 11.06 & 1.67 & 9.67 && 16.91 &  5.55 &  15.58 & 29.2 \\  
    \textsc{Prefix}       & 22.43 & 6.49 & 19.65 &&  23.14 & 8.25 & 21.73 & 100 \\
    \textsc{Compress}     & 19.77 & 4.02 & 17.30 && 19.63 & 5.13 & 18.28   &  100 \\
    \textsc{W\&L}         & 22 & 6 & 17 && -   & -   & -   & -   \\
    \textsc{Topiary}      & 25.12 & 6.46 & 20.12 && -   & -   & -   & -  \\

    \textsc{Moses+}       & 26.50 & 8.13 & 22.85 && 28.77 & 12.10 & 26.44 & 70.5 \\
    \textsc{Abs}          & 26.55 & 7.06 & 22.05 &&  30.88& 12.22 & 27.77 & 85.4 \\
    \textsc{Abs+}         & 28.18 & 8.49 & 23.81 &&  31.00 & 12.65 & 28.34 & 91.5 \\ 
    \midrule

    \textsc{Reference}    & 29.21 & 8.38 & 24.46  &&  -  & -   & -   &  45.6 \\
    \bottomrule
  \end{tabular}
  
  \caption{\small \label{tab:results} Experimental results on the main summary tasks on various ROUGE metrics . Baseline models are described in detail in 
    Section~\ref{sec:baseline}. We report the percentage of tokens in the summary that also appear in the input for Gigaword as \texttt{Ext \%}.}
\end{table*}

\subsection{Baselines}
\label{sec:baseline}

Due to the variety of approaches to the sentence summarization
problem, we report a broad set of headline-generation baselines.

From the DUC-2004 task we include the \textsc{Prefix} baseline
that simply returns the first 75-characters of the input as the
headline.  We also report the winning system on this shared task,
\textsc{Topiary} \cite{zajic2004bbn}. \textsc{Topiary} merges a
compression system using linguistically-motivated transformations of
the input \cite{dorr2003hedge} with an unsupervised topic detection
(UTD) algorithm that appends key phrases from the full article onto
the compressed output. \newcite{woodsend2010generation} (described above) also report results on the
DUC dataset.

The DUC task also includes a set of manual summaries performed by 8
human summarizers each summarizing half of the test data sentences (yielding 4 references per sentence). We report the
average inter-annotater agreement score as \textsc{Reference}. For reference, the best human evaluator scores 31.7 ROUGE-1.

We also include several baselines that have access to the same
training data as our system. The first is a sentence compression
baseline \textsc{Compress} \cite{clarke2008global}. This model uses
the syntactic structure of the original sentence along with a language
model trained on the headline data to produce a compressed output.  The syntax and language model
are combined with a set of linguistic constraints and decoding is performed with an ILP solver.

To control for memorizing titles from training, we implement an
information retrieval baseline, \textsc{IR}. This baseline indexes the
training set, and gives the title for the article with highest
BM-25 match to the input (see \newcite{manning2008introduction}).

Finally, we use a   phrase-based statistical machine translation system trained on Gigaword
to produce summaries, \textsc{Moses+} \cite{koehn2007moses}. To improve the baseline for this task, we augment the phrase table
with ``deletion'' rules mapping each article word to $\epsilon$,
include an additional deletion feature for these rules, and allow for an
infinite distortion limit. We also explicitly tune the model using
MERT to target the 75-byte capped ROUGE score as opposed to standard
BLEU-based tuning. Unfortunately, one remaining issue is that it is non-trivial to modify the translation decoder to produce fixed-length outputs, so we tune the system to produce roughly the expected length.

\subsection{Implementation}

For training, we use mini-batch stochastic gradient descent to
minimize negative log-likelihood. We use a learning rate of $0.05$,
and split the learning rate by half if validation log-likelihood does
not improve for an epoch. Training is performed with shuffled
mini-batches of size 64. The minibatches are grouped by input
length. After each epoch, we renormalize the embedding tables \cite{hinton2012improving}. Based on the
validation set, we set hyperparameters as $D=200$, $H=400$, $C=5$,
$L=3$, and $Q = 2$.

Our implementation uses the Torch numerical framework
(\url{http://torch.ch/}) and will be openly available along with the
data pipeline. Crucially, training is performed on GPUs and would be
intractable or require approximations otherwise. Processing 1000
mini-batches with $D=200$, $H=400$ requires 160 seconds.  Best
validation accuracy is reached after 15 epochs through the data, which
requires around 4 days of training.

Additionally, as described in Section~\ref{sec:tuning} we apply a
MERT tuning step after training using the DUC-2003 data.  For this step we use Z-MERT
\cite{zaidan2009z}. We refer to the
main model as \textsc{Abs} and the tuned model as
\textsc{Abs+}.

%% file: experiments.tex
\section{Results}
\label{sec:results}

Our main results are presented in Table~\ref{tab:results}. We run
experiments both using the DUC-2004 evaluation data set (500
sentences, 4 references, 75 bytes) with all systems and a randomly
held-out Gigaword test set (2000 sentences, 1 reference). We first
note that the baselines \textsc{Compress} and \textsc{IR} do
relatively poorly on both datasets, indicating that neither just
having article information or language model information alone is
sufficient for the task. The \textsc{Prefix} baseline actually
performs surprisingly well on ROUGE-1 which makes sense given the
earlier observed overlap between article and summary.

Both \textsc{Abs} and \textsc{Moses+} perform better than
\textsc{Topiary}, particularly on ROUGE-2 and ROUGE-L in DUC.  The
full model \textsc{Abs+} scores the best on these tasks, and is
significantly better based on the default ROUGE confidence level than
\textsc{Topiary} on all metrics, and \textsc{Moses+} on
ROUGE-1 for DUC as well as ROUGE-1 and ROUGE-L for Gigaword. Note that the additional extractive features bias the
system towards retaining more input words, which is useful for the
underlying metric.

Next we consider ablations to the model and algorithm
structure. Table~\ref{tab:perp} shows experiments for the model with
various encoders. For these experiments we look at the perplexity of
the system as a language model on validation data, which controls for
the variable of inference and tuning.  The NNLM language model with no
encoder gives a gain over the standard n-gram language
model. Including even the bag-of-words encoder reduces perplexity
number to below 50. Both the convolutional encoder and the
attention-based encoder further reduce the perplexity, with
attention giving a value below 30.

We also consider model and decoding ablations on the main summary
model, shown in Table~\ref{tab:ablations}. These experiments
compare to the BoW encoding models, compare beam search and greedy
decoding, as well as restricting the system to be complete
extractive. Of these features, the biggest impact is from using a more
powerful encoder (attention versus BoW), as well as using beam search
to generate summaries. The abstractive nature of the system helps, but
for ROUGE even using pure extractive generation is effective.

\begin{table}
  \centering
  \small
  \begin{tabular}{lcc}
    \toprule
    Model & Encoder & Perplexity  \\
    \midrule
    KN-Smoothed 5-Gram  & none   & 183.2  \\
    Feed-Forward NNLM & none    & 145.9  \\
    Bag-of-Word   & enc$_1$  & 43.6   \\
    Convolutional (TDNN)  & enc$_2$  & 35.9  \\
    Attention-Based (\textsc{Abs})  & enc$_3$  & 27.1 \\
    \bottomrule
  \end{tabular}
  \caption{\small \label{tab:perp} Perplexity results on the Gigaword validation set comparing various language models with C=5 and end-to-end summarization models. The encoders are defined in Section~\ref{sec:model}.
  }
\end{table}

\begin{table}
  \centering
  \small
  \begin{tabular}{lllccc}
    \toprule
    Decoder &  Model & Cons.     & R-1 & R-2 & R-L \\
    \midrule 
    Greedy & \textsc{Abs+} & Abs & 26.67& 6.72& 21.70 \\
    Beam & \textsc{BoW}   & Abs & 22.15& 4.60& 18.23 \\ 
    Beam & \textsc{Abs+}   & Ext & 27.89& 7.56& 22.84 \\ 
    Beam & \textsc{Abs+} & Abs & 28.48& 8.91& 23.97 \\ 
    \bottomrule
  \end{tabular}
  \caption{\small \label{tab:ablations} ROUGE scores on DUC-2003 development data for various versions of inference. Greedy and Beam are described in Section~\ref{sec:additional}. Ext. is a purely extractive version of the system (Eq. \ref{eq:ext}) \vspace{-0.4cm}}
\end{table}
Finally we consider example summaries shown in
Figure~\ref{fig:examples}.  Despite improving on the baseline scores,
this model is far from human performance on this task.  Generally the
models are good at picking out key words from the input, such as names and
places. However, both models will reorder words in syntactically
incorrect ways, for instance in Sentence~7 both models have the
wrong subject. \textsc{Abs} often uses 
more interesting re-wording, for instance \textit{new nz pm after election} 
in Sentence~4, but this can also lead to attachment mistakes such a \textit{russian oil giant chevron}
in Sentence~11.

\begin{figure}
  \scriptsize
  \begin{framed}    

\textbf{I(1):} a detained iranian-american academic accused of acting against national security has been released from a tehran prison after a hefty bail was posted , a to
p judiciary official said tuesday . \\
\textbf{G:} iranian-american academic held in tehran released on bail \\
\textbf{A:} detained iranian-american academic released from jail after posting bail \\
\textbf{A+:} detained iranian-american academic released from prison after hefty bail \\

\textbf{I(2):} ministers from the european union and its mediterranean neighbors gathered here under heavy security on monday for an unprecedented conference on economic and political cooperation . \\
\textbf{G:} european mediterranean ministers gather for landmark conference by julie bradford \\
\textbf{A:} mediterranean neighbors gather for unprecedented conference on heavy security \\
\textbf{A+:} mediterranean neighbors gather under heavy security for unprecedented conference \\

\textbf{I(3):} the death toll from a school collapse in a haitian shanty-town rose to \#\# after rescue workers uncovered a classroom with \#\# dead students and their teacher , officials said saturday . \\
\textbf{G:} toll rises to \#\# in haiti school unk : official \\
\textbf{A:} death toll in haiti school accident rises to \#\# \\
\textbf{A+:} death toll in haiti school to \#\# dead students \\

\textbf{I(4):} australian foreign minister stephen smith sunday congratulated new zealand 's new prime minister-elect john key as he praised ousted leader helen clark as a `` gutsy '' and respected politician . \\
\textbf{G:} time caught up with nz 's gutsy clark says australian fm \\
\textbf{A:} australian foreign minister congratulates new nz pm after election \\
\textbf{A+:} australian foreign minister congratulates smith new zealand as leader \\

\textbf{I(5):} two drunken south african fans hurled racist abuse at the country 's rugby sevens coach after the team were eliminated from the weekend 's hong kong tournament , reports said tuesday . \\
\textbf{G:} rugby union : racist taunts mar hong kong sevens : report \\
\textbf{A:} south african fans hurl racist taunts at rugby sevens \\
\textbf{A+:} south african fans racist abuse at rugby sevens tournament \\

\textbf{I(6):} christian conservatives -- kingmakers in the last two us presidential elections -- may have less success in getting their pick elected in \#\#\#\# , political observers say . \\
\textbf{G:} christian conservatives power diminished ahead of \#\#\#\# vote \\
\textbf{A:} christian conservatives may have less success in \#\#\#\# election \\
\textbf{A+:} christian conservatives in the last two us presidential elections \\

\textbf{I(7):} the white house on thursday warned iran of possible new sanctions after the un nuclear watchdog reported that tehran had begun sensitive nuclear work at a key site in defiance of un resolutions . \\
\textbf{G:} us warns iran of step backward on nuclear issue \\
\textbf{A:} iran warns of possible new sanctions on nuclear work \\
\textbf{A+:} un nuclear watchdog warns iran of possible new sanctions \\

\textbf{I(8):} thousands of kashmiris chanting pro-pakistan slogans on sunday attended a rally to welcome back a hardline separatist leader who underwent cancer treatment in mumbai . \\
\textbf{G:} thousands attend rally for kashmir hardliner \\
\textbf{A:} thousands rally in support of hardline kashmiri separatist leader \\
\textbf{A+:} thousands of kashmiris rally to welcome back cancer treatment \\

\textbf{I(9):} an explosion in iraq 's restive northeastern province of diyala killed two us soldiers and wounded two more , the military reported monday . \\
\textbf{G:} two us soldiers killed in iraq blast december toll \#\#\# \\
\textbf{A:} \# us two soldiers killed in restive northeast province \\
\textbf{A+:} explosion in restive northeastern province kills two us soldiers \\

\textbf{I(10):} russian world no. \# nikolay davydenko became the fifth withdrawal through injury or illness at the sydney international wednesday , retiring from his second round match with a foot injury . \\
\textbf{G:} tennis : davydenko pulls out of sydney with injury \\
\textbf{A:} davydenko pulls out of sydney international with foot injury \\
\textbf{A+:} russian world no. \# davydenko retires at sydney international \\

\textbf{I(11):} russia 's gas and oil giant gazprom and us oil major chevron have set up a joint venture based in resource-rich northwestern siberia , the interfax news agency reported thursday quoting gazprom officials . \\
\textbf{G:} gazprom chevron set up joint venture \\
\textbf{A:} russian oil giant chevron set up siberia joint venture \\
\textbf{A+:} russia 's gazprom set up joint venture in siberia 

  \end{framed}
  \vspace*{-0.5cm}
  \caption{\small \label{fig:examples} Example sentence summaries produced on Gigaword. \textbf{I} is the input, \textbf{A} is \textsc{Abs}, and \textbf{G} is the true headline.}
\end{figure}

\section{Conclusion}

We have presented a neural attention-based model for abstractive
summarization, based on recent developments in neural machine
translation. We combine this probabilistic model with a generation
algorithm which produces accurate abstractive summaries. As a next
step we would like to further improve the grammaticality of the
summaries in a data-driven way, as well as scale this system to
generate paragraph-level summaries. Both pose additional challenges in
terms of efficient alignment and consistency in
generation.